# Guidelines and Benchmarks for Deployment of Deep Learning Models on Smartphones as Real-Time Apps

Abhishek Sehgal and Nasser Kehtarnavaz

*Abstract*— Deep learning solutions are being increasingly used in mobile applications. Although there are many open-source software tools for the development of deep learning solutions, there are no guidelines in one place in a unified manner for using these tools towards real-time deployment of these solutions on smartphones. From the variety of available deep learning tools, the most suited ones are used in this paper to enable real-time deployment of deep learning inference networks on smartphones. A uniform flow of implementation is devised for both Android and iOS smartphones. The advantage of using multi-threading to achieve or improve real-time throughputs is also showcased. A benchmarking framework consisting of accuracy, CPU/GPU consumption and real-time throughput is considered for validation purposes. The developed deployment approach allows deep learning models to be turned into real-time smartphone apps with ease based on publicly available deep learning and smartphone software tools. This approach is applied to six popular or representative convolutional neural network models and the validation results based on the benchmarking metrics are reported.

## I. INTRODUCTION

Deep learning has had a dramatic impact on advancing the field of machine learning [1]. It has pushed the state-of-the-art beyond what conventional approaches have achieved in various applications such as object detection [2], object localization [3], and speech recognition [4]. The expansion in the use of deep learning has been fueled by increases in the computational power of processors, in particular graphics processing units (GPUs), and the availability of large datasets for training.

Deep learning involves Deep Neural Networks (DNNs) consisting of a cascade of non-linear processing units arranged in layers by which an increasing level of data abstraction is enabled at deeper layers. This is of particular importance in classification and regression tasks due to the fact that raw or minimally processed data can get processed without the need to perform feature extraction as compared to conventional approaches that normally require obtaining hand-crafted features first. DNNs are able to learn optimal features themselves for a particular task and have provided state-of-the-art accuracies in computer vision, speech recognition, natural language processing applications among others.

In terms of implementation platforms, smartphones have emerged as a ubiquitous and mobile computing device with more than 2.5 billion people worldwide owning them [5]. Apart from being equipped with multi-core CPUs and GPUs, smartphones contain a plethora of sensors which do not require interfacing hardware as compared to other popular platforms such as Arduino [6] and Raspberry Pi [7]. In addition, there exist well developed and supported Application Programming Interfaces (APIs) for smartphones which have been optimized for performance.

Smartphones constitute the highest users of deep learning-based solutions spanning various applications such as voice assistants, automatic text prediction, and augmented reality. Also, they are used as research platforms to run deep learning solutions involving different applications such as concussion detection [8], jaundice diagnosis [9], schizophrenia recognition [10] and voice activity detection for hearing studies [11] among others. These solutions either rely on server-side processing or perform offline simulations on the data that are previously collected using smartphones. Real-time deployment of such deep learning solutions on smartphones has been fairly limited in the literature. Manual coding of DNNs to run in real-time on smartphones is cumbersome and time consuming. This paper makes the process of deployment of deep learning algorithms on smartphones easy by providing in one place the steps needed to bridge the gap between development and deployment based on the publicly available software tools.

There have been some works in the literature on the development of deep learning solutions that are aimed specifically at mobile implementation [12],[13],[14],[15]. On-device deep learning engines are also finding their way into smartphones. For example, Apple has introduced a neural engine as part of the A11 Bionic chip and Huawei has introduced the Kirin 970 neural processing unit (NPU). The smartphone industry is also working toward dedicated processors to speed up on-device deep learning in contrast to cloud servers in order to cope with real-time implementation issues and the need for internet connection. In addition, on-

A. Sehgal and N. Kehtarnavaz are with the Department of Electrical and Computer Engineering, University of Texas at Dallas, Richardson, TX 75080, USA. E-mail: {abhishek.sehgal, kehtar}@utdallas.edu.

device deep learning helps to alleviate security or privacy concerns due to data storage on servers. Considering that smartphones are equipped with multi-core CPUs, multithreading is used here to reduce computation time towards achieving real-time throughputs.

Furthermore, the open-source deep learning software tools have reached a maturation point in terms of libraries for on-device deep learning deployment. However, there exists a steep learning curve associated with the deployment of these software tools and an absence of benchmarking guidelines for smartphones. Although previous works have addressed efficient processing techniques for the purpose of running DNN models on smartphones, thus far no step-by-step guidelines or benchmarks have been provided regarding the real-time deployment of DNN models on smartphones. This paper aims at bringing such information into one place or under one umbrella, thus providing a unified approach to easily deploy trained deep learning models as apps on Android and iOS smartphones with a focus on their real-time operation. This work enables smartphones to be used as a portable research platform for deep learning studies.

Towards this objective, the rest of the paper is organized as follows: Section II describes the most suited deep learning libraries for smartphone deployment at the time of this writing, deployment steps based on the smartphone operating system, the software tools used to build deep learning apps, and the smartphone devices used to showcase a number of representative deep learning models. In Section III, the DNN models and the benchmarking criteria used for validation are discussed. The use of multi-core CPUs on the smartphones to achieve or improve real-time throughputs through multi-threading is also discussed in this section. Section IV provides the validation results and their discussion. Finally, the paper is concluded in Section V.

## II. Deployment of DNN Models on Smartphones

This section discusses how to deploy DNN models on smartphones using publicly and freely available software tools. The steps discussed is aimed at turning DNN models into apps for both Android and iOS smartphones in a unified manner.

### A. Deep Learning Software Tools

The rise of deep learning has been accelerated by the introduction of various publicly and freely available libraries. The main libraries that are widely used include Caffe [16] (developed by Berkeley AI Research), TensorFlow [17] (developed by Google), PyTorch [18] (developed by Facebook), and CNTK [19] (developed by Microsoft). These libraries support Python for the purpose of training and prototyping models. Even though such a wide collection of prototyping tools is available to train and develop deep learning models, researchers often wonder where to begin. Here, three publicly available libraries most suited for the task under consideration are selected from the available deep learning libraries. These libraries are selected based on (i) their easy portability to mobile devices and (ii) active support by their developers. In what follows, these libraries and their frameworks are briefly described.

*TensorFlow* is a dataflow programming library. It expresses computations as stateful dataflow graphs, enabling users to define a neural network as a graph of operations that can be executed on input data streams. Data are represented as multidimensional arrays or "tensors" thus the name TensorFlow. The underlying benefit of defining a neural network as a graph is that the computations and memory usage are highly optimized and they can be parallelized using multiple CPUs and GPUs and implemented across a variety of hardware platforms. As TensorFlow and Android are both developed by Google, TensorFlow models can be integrated into the Android software environment with ease by adding the "TensorFlow for Mobile" library module as a dependency. A similar tool named "TensorFlow Lite" is also available, but currently it remains experimental and does not support as many operations as TensorFlow for Mobile. Furthermore, it is worth noting that the deployment flow for TensorFlow Lite is similar to TensorFlow for Mobile.

Another widely used library is called *Keras* [20], which is a higher-level library written in Python where TensorFlow or CNTK can be used as its backend. Keras makes development of models easier and faster by providing the building blocks for common-use DNN layers, a simple coding syntax, and tools to easily preprocess data. As Keras can use TensorFlow as its backend, the model trained using Keras is essentially a TensorFlow model which can be extracted and used in Android apps.

*CoreML* [21] is a software framework developed by Apple to run machine learning models on iOS devices. It has a Python-based tool called CoreMLTools [22] which allows one to translate existing machine learning models into CoreML supported models. This conversion capability allows the conversion of Keras models into CoreML models which can then be implemented as an app on iOS devices or iPhones. A converter developed by TensorFlow (tf-coreml) [23] also allows converting TensorFlow models into CoreML models.

The latest version of CoreMLTools at the time of this writing is 0.8, which supports Keras version 2.1.3 and TensorFlow version 1.5. These versions are utilized here for the results reported in the paper.

### B. Deployment Steps

The steps that are needed for deployment of deep learning models on smartphones are showcased in Fig. 1. Keras can be considered to be the primary prototyping library as it can be easily converted to CoreML models for iOS and the underlying TensorFlow backend model can be extracted from it for Android. In case of TensorFlow models, a secondary path (marked in red in Fig. 1) is also provided by using the converter tf-coreml to convert the models into CoreML for implementation on iOS smartphones. In case of models trained using different libraries, several publicly available converter tools [24] are available to convert them to Keras or TensorFlow. CoreML also provides conversion tools for models trained by the deep-learning libraries other than Keras.

Models are stored as inference only and all training related layers are removed to allow only the feed-forward path of a network model to execute. Since layers are stored as computational graphs, this allows their optimization for the platform they are going to run on. The flowcharts provided in

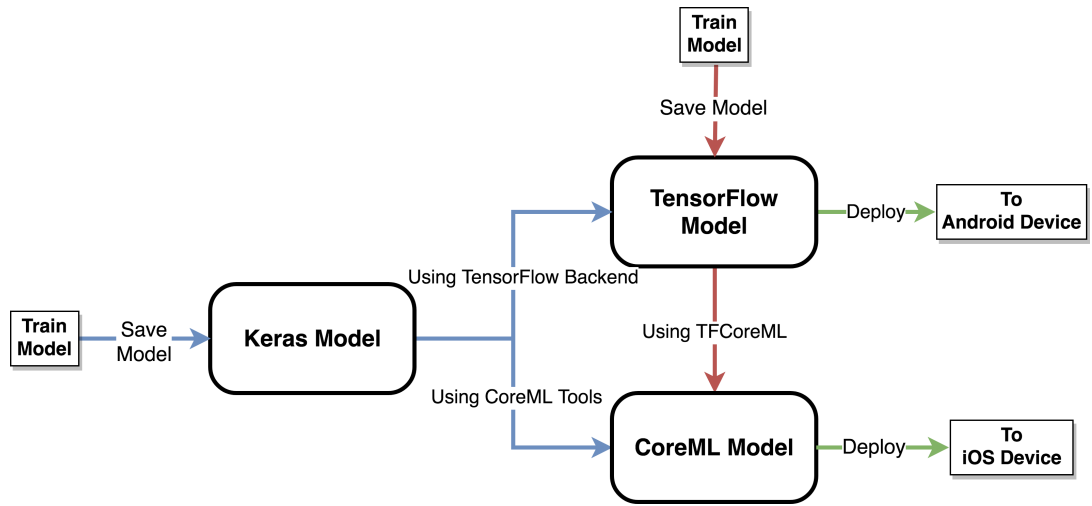

Fig. 1. Diagram illustrating the utilization of publicly available software tools for deployment of deep learning solutions on smartphones – the blue lines illustrate the path when Keras is used as the primary framework for training deep learning models and the red lines illustrate the path when TensorFlow is used as the primary framework for training deep learning models.

Figs. 2 and 3 depict the steps needed to convert a trained Keras or TensorFlow model to a model for deployment on Android and iOS smartphones. The flowcharts appearing in Fig. 4 show the steps needed to create an Android or iOS app from a converted model.

*1) Model Generation – iOS*

To create a CoreML model, a Keras model needs to be trained first or a pre-trained model needs to be considered first. Keras models are usually saved as a .h5 file which denotes the Hierarchical Data Format (HDF). This allows using the same trained model across different backends. This file format stores the architecture of the graph and the weights of the graph tensors as numerical arrays. After the model is loaded into Python, the CoreMLTools python library can be used to convert the Keras model to a CoreML model. The converter provides the option to specify the input as an image or as a multi-dimensional array. In case of image inputs, the converter provides the option to define the pre-processing parameters used for that model. As the pre-processing varies across different models, this option is highly useful since it allows a model to be used on raw-images and easy switching without explicitly implementing pre-processing for each different model.

The converted CoreML model is stored as a .mlmodel file. This file encapsulates a MLModel class which can be directly instantiated allowing the model to be used as a plug-and-play model. The CoreML API handles all the underlying DNN computation removing the overhead required for coding a neural network from scratch. As a result, the user can focus on deploying and testing the model rather than implementing the neural network.

*2) Model Generation – Android*

TensorFlow for Mobile provides a Gradle build dependency [25]. This allows using predefined functions for inference via just a trained model. A TensorFlow model gets stored as a .pb file which denotes the Protocol Buffer file format. Similar to HDF, this format also stores the architecture of a model and its trained weights. The TensorFlow for Mobile inference interface creates a model based on a .pb file which can then be executed on Android smartphones using the predefined functions included in the dependency.

To extract a TensorFlow model from a Keras model, first the variables in the model need to be converted to constants. Variable tensors are only required for training as they get updated based on the back-propagation input. For inference only models, these tensors need to be constant. This can be done by the graph utility sub-module (graph_util) in TensorFlow. To convert a graph, first the underlying TensorFlow session created by Keras needs to be accessed. This can be achieved by using the backend module in Keras. Then, the session graph can be inputted to the graph_util constant converter function to obtain the graph with constant weights. This constant graph can be saved as a .pb file using the graph input/output sub-module (graph_io), which can get imported into an Android app using the TensorFlow inference interface. This interface creates a session similar to the TensorFlow session in Python and handles all the required DNN computations. The input image or data can then be fed into the session to extract the output of the model from the session. Compared with CoreML, one needs to explicitly pre-process the image in this method. However, an advantage that TensorFlow for Mobile has is that the output of any of the intermediate layers can be extracted, whereas in CoreML only the output of the model can be extracted. Unlike CoreML, one needs to manually set up the model by explicitly feeding the .pb file, the input and output node names, and the size of the input.

Treating TensorFlow as the primary framework for DNN development, the converter tf-coreml can be used to convert a .pb model into a .mlmodel file. This converter tool provides a reduced set of computations and this set can be seen on the GitHub page of the converter for the purpose of altering the model if any unsupported computations are seen.

*C. Smartphone Software Tools*

To demonstrate on-device deep learning inference, both Android and iOS smartphones are considered in this work to

4form a unified approach. These two operating systems have a combined market share of 96% of smartphones worldwide [26]. Additionally, the developer tools for both of these smartphones operating systems are available online for free and are well maintained by their respective organizations.

To develop Android apps, the Android Studio IDE [27] is used which is available for all operating systems. The language of choice for Android development is Java, which is used here to develop apps to run DNN models on Android smartphones. Android apps can also be packaged as executable Android Application Package (APK) files for deployment on any Android smartphone.

iOS apps can be developed and deployed on an iOS device or iPhone only via a macOS machine running the Xcode IDE [28]. iOS apps are developed using the Swift or Objective-C programming language. To deploy iOS apps on an iPhone, one needs to be registered as an Apple Developer.

*D. Smartphone Processors*

For running DNN models, two modern smartphones of Pixel 2 and iPhone 8 are used in this work as sample Android and iOS smartphones, respectively.

Pixel 2 is developed by Google. It runs Android operating system and possesses a Qualcomm Snapdragon 835 64-bit ARM-based octa-core system on a chip (SoC). Its CPU clock speed varies between 1.9-2.35 GHz depending on the core being used. The internal memory of this smartphone is 4 GB LPDDR4x RAM. It also possesses an Adreno 540 GPU. Note that TensorFlow for Mobile does not utilize this GPU. The Pixel 2 smartphone used here runs the latest Android version 8.1.0 at the time of this writing.

On the iOS side, iPhone 8 is used here which incorporates the Apple A11 Bionic 64-bit ARM-based hexa-core SoC with a maximum CPU clock rate of 2.39 GHz. The internal memory of iPhone 8 is 2 GB of LPDDR4x RAM. The A11 chip also contains a dedicated neural engine which can be used to run machine learning models more efficiently than using plain GPU. The neural engine is capable of performing up to 6000 billion operations per second. The iPhone 8 smartphone used here runs the latest iOS version 11.4 at the time of this writing.

## III. DNN SMARTPHONE APPS AND BENCHMARKING METRICS

*A. DNN Models*

In this section, the steps involved in turning DNN models to smartphone apps are applied to six popular Convolutional Neural Networks (CNNs) and a benchmarking framework of these models is discussed. CNNs are a class of DNNs where the primary computation involves convolution. The convolution layers (CONV) in CNNs are able to provide a higher level of abstraction by creating feature extraction kernels similar to those used in image processing. By stacking convolution layers, a CNN model is able to extract unique information related to a particular image or matrix input. CNN models currently provide the state-of-the-art solutions in many image processing, computer vision, speech and audio processing applications. In this paper, six popular or representative CNNs are benchmarked based on the MNIST [29] dataset, which is considered to be a gateway dataset for exploring deep learning, and the widely used ImageNet Large Scale Visual Recognition Competition (ILSVRC) [30] dataset. These networks are briefly described below for the sake of completeness.

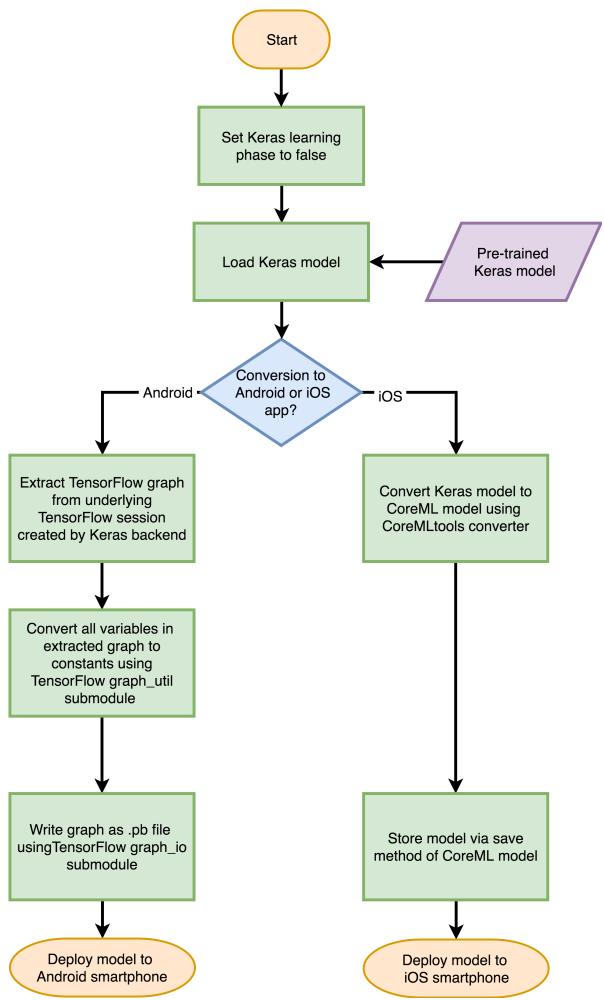

Fig. 2. Flowchart depicting the steps needed to convert a Keras model into a smartphone deployable model for Android and iOS smartphones.

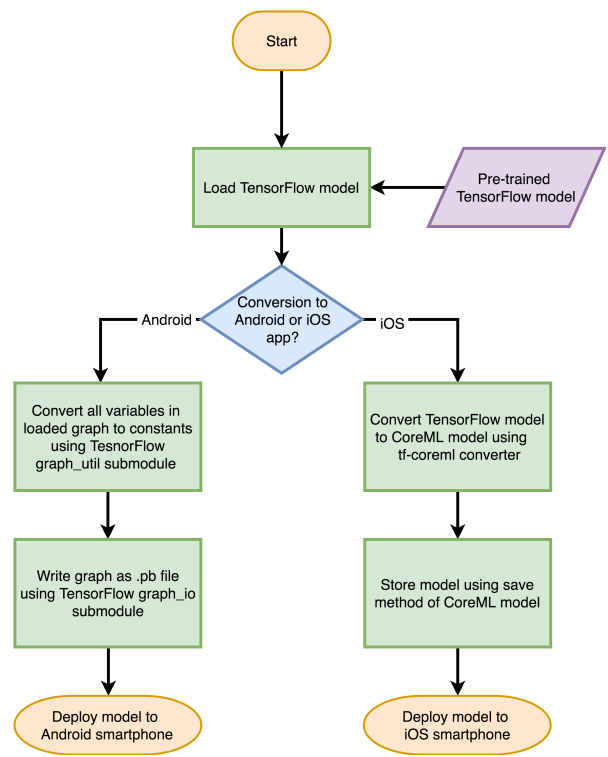

Fig. 3. Flowchart depicting the steps needed to convert a TensorFlow model into a smartphone deployable model for Android and iOS smartphones.

*LeNet* was introduced in [31] for digit classification trained on the MNIST dataset. The summary of the model is provided in Table I. It is designed to classify grayscale images of dimensions $28 \times 28$ into single digits using 2 CONV layers followed by 3 fully-connected (FC) layers. To reduce the dimensions of the intermediate feature maps, $2 \times 2$ average pooling is utilized. The activation function used in LeNet is sigmoid.

*ResNet* [32] or Residual Network is a CNN model that include so called "skip" or "shortcut" connections which allow bypassing the weight layers using identity mappings. The output of the weight layers and the identity mapping are then added together. The skip connections of ResNet are critical in preventing the gradient from vanishing in deep layers, as the identity mapping prevents the backpropagation error from shrinking. The weight layers in ResNet usually consist of two $3 \times 3$ CONV layers. To reduce the number of parameters in each weight layer, ResNet also uses so called "bottleneck" layers by using $1 \times 1$ filters. The bottleneck layers replace the two layers with three layers of $1 \times 1$, $3 \times 3$ and $1 \times 1$ filters. The $1 \times 1$ filters are used to decrease and then increase the number of weights. ResNet-152 was selected as the winner of the ILSVRC 2015 challenge, surpassing human level accuracy with a top-5 accuracy of 3.57% on the test set provided. Here, the ResNet-50 model is used which has 1% less accuracy than ResNet-152, but with 2.5 times fewer parameters (approx. 25.6 million). It consists of a CONV layer followed by 16 bottleneck layers and a FC layer.

*InceptionV3* [33] is the extension of *GoogLeNet* [34] that was selected as the winner of the ILSVRC 2014 challenge. GoogLeNet is based on the inception module which consists of 4 parallel CONV layers of $1 \times 1$ CONV, $3 \times 3$ and $1 \times 1$ CONV, $5 \times 5$ and $1 \times 1$ CONV, and $1 \times 1$ CONV followed by $3 \times 3$ max-pooling. GoogLeNet achieved a top-5 error of 6.65% on the ILSVRC challenge test set. InceptionV3 includes the following 3 new inception modules: (i) In the first module, the $5 \times 5$ CONV is replaced with two $3 \times 3$ CONV to reduce the number of parameters in the module. (ii) In the second module, the n×n CONV layer is replaced with an n×1 CONV followed by a 1×n. This reduces the number of weights and thus the computational cost. For the results reported in the next section, n is considered to be 7. (iii) The third module separates the initial $3 \times 3$ CONV in the first inception module into two parallel $3 \times 1$ and $1 \times 3$ CONV layers. This module is used to promote high dimensional sparse representations and is placed last after the other two modules.

InceptionV3 consists of 6 CONV layers followed by 3 of the first modified inception modules, 5 of the second modified inception modules, 2 of the third modified inception modules and a FC layer. It also contains an auxiliary classifier [34], which is an inception module on the output of the second modified inception module with a batch-normalized [35] FC layer to increase accuracy. InceptionV3 achieves a top-5 accuracy similar to ResNet-152 while having only 23.8 million

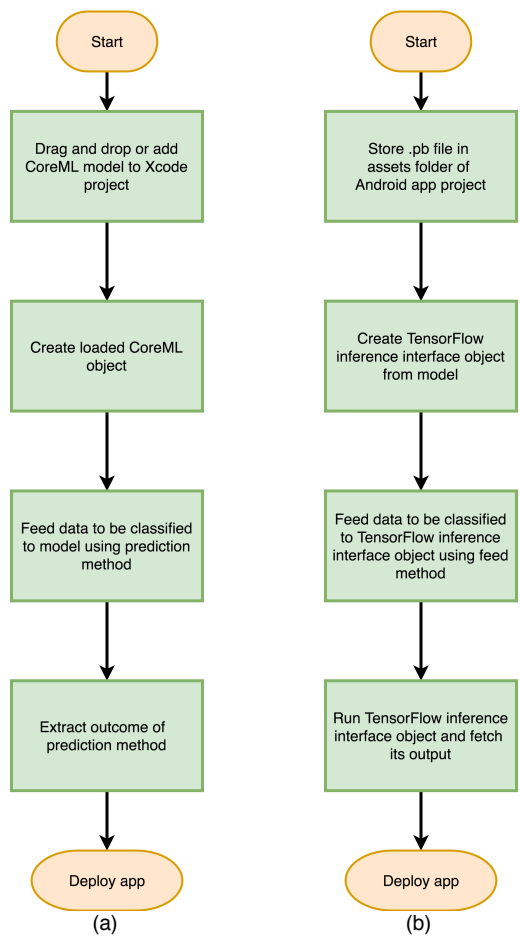

Fig. 4. Flowcharts depicting the steps needed for creating an (a) Android, and (b) iOS app from a converted model.

TABLE I
SUMMARY OF LENET MODEL TRAINED ON MNIST DATASET

| Layer Information | LeNet |
| --- | --- |
| **Input-Size** | 28x28x1 |
| **CONV Layer** | |
| # of CONV Layers | 2 |
| Depth | 2 |
| Kernel Size | 5 |
| Strides | 1 |
| # of Channels | 1, 16 |
| # of Filters | 16, 32 |
| **FC Layer** | |
| # of FC Layers | 3 |
| # of Channels | 128 - 3200 |
| # of Filters | 10 - 256 |
| **Parameters** | 866K |
| **FLOPs** | 28M |
| **Model Storage Memory** | 3.5MB |
| **Accuracy** | 99.43 |

TABLE II
SUMMARY OF POPULAR CNNS TRAINED ON THE ILSVRC CHALLENGE DATASET

| Layer Information | ResNet50 | InceptionV3 | SqueezeNet | MobileNet | DenseNet |
|---|---|---|---|---|---|
| **Input-Size** | 224x224x3 | 299x299x3 | 227x227x3 | 224x224x3 | 224x224x3 |
| **CONV Layer** | | | | | |
| # of CONV Layers | 49 | 95 | 26 | 27 | 120 |
| Depth | 49 | 46 | 18 | 27 | 120 |
| Kernel Size | 1,3,7 | 1,3,5,7 | 1,3 | 1,3 | 1,3,7 |
| Strides | 1,2 | 1,2 | 1,2 | 1,2 | 1,2 |
| # of Channels | 3 - 2048 | 3 - 2048 | 3 - 1000 | 3 - 1024 | 3 - 1024 |
| # of Filters | 64 - 2048 | 32 - 2048 | 16 - 1000 | 32 - 1024 | 32 - 1024 |
| **FC Layer** | | | | | |
| # of FC Layers | 1 | 1 | 0 | 1 | 1 |
| # of Channels | 2048 | 2048 | 0 | 1024 | 1024 |
| # of Filters | 1000 | 1000 | 0 | 1000 | 1000 |
| **Parameters** | 25.6 M | 23.8 M | 1.2 M | 4.3 M | 8 M |
| **FLOPs** | 7.7 B | 11.5 B | 714 M | 1.1 B | 5.7 B |
| **Model Storage Memory** | 102 MB | 96 MB | 5 MB | 17MB | 33MB |
| **Top-5 Accuracy (Single Crop)** | 92.1% | 93.8% | 78.4% | 86.2% | 91.8% |

parameters (similar to ResNet-50).

*SqueezeNet* [36] is a CNN model designed for limited-memory systems. It provides the accuracy of *AlexNet* [2]. This model was selected as the winner of the ILSVRC challenge 2012 with a top-5 error of 9.8% with 50 times fewer parameters. SqueezeNet is built using "fire" modules which consist of two stacked layers: squeeze layer, and expand layer. The squeeze layer is composed exclusively of $1 \times 1$ filters, which reduce the number of channels of the input to the module. The expand layer is a mix of $1 \times 1$ and $3 \times 3$ filters, the outputs of which are concatenated after activation and fed into the next module. The sparing use of $1 \times 1$ reduces the number of parameters of the CNN model considerably, while still maintaining the baseline accuracy. SqueezeNet has been compressed even further using Deep Compression [15], with the reduction in size of SqueezeNet being 510 times that of AlexNet with no loss in accuracy. The downside to this is that the compressed model cannot be used using the existing deep learning software tools. SqueezeNet consists of a CONV layer followed by 8 so-called fire modules and a CONV layer in the end. FC layers are not used as they have much higher number of parameters as compared to CONV layers.

*MobileNet* [14] is a CNN model that has been specifically designed for mobile and embedded vision applications. MobileNet modules reduce computations and memory by dividing a normal CONV layer into two parts: depthwise convolution and pointwise convolution. These two parts together are called Depthwise Separable Convolution. In depthwise convolution, the channel width of the filter is kept as 1. The pointwise convolution uses $1 \times 1$ filters to expand the channels of the output of the depthwise convolution. MobileNets can be modified by using width and resolution multipliers, called model shrinking hyperparameters. The width multiplier is used to reduce the channels of the network uniformly at each layer, thereby reducing the overall number of parameters of the layer. The resolution multiplier is implicitly set by changing the input resolution of the model, which reduces the computational cost of the model. Here, MobileNet is considered with a width multiplier of 1.0 and an input resolution of $224 \times 224$. It consists of a CONV layer followed by 13 depthwise separable convolution layers and a FC layer.

*DenseNet* [37] or Densely Connected Networks is a CNN that extends the residual learning framework introduced by ResNets. In a DenseNet block, every layer is connected to all the subsequent layers of equal feature map dimensions. Instead of additions, as done in ResNet, the input is concatenated. The CONV architecture used in DenseNet is similar to the bottleneck architecture in ResNet, where a $1 \times 1$ filter is used to reduce the number of the channels of the input before feeding it into the $3 \times 3$ CONV layer. Between every DenseNet block, a compression/transition layer is used to reduce the number of feature maps into the next DenseNet block. DenseNet requires considerably fewer number of parameters than ResNet to achieve similar accuracy. Here, DenseNet-121 is used, which gives 1% less accuracy than ResNet50 with 3 times fewer parameters. It consists of a CONV layer followed by 4 DenseNet Blocks with 3 transition layers between them followed by a FC layer.

Table II shows a summary of the CNN models trained on the ILSVRC challenge dataset. A comparison of the models based on their top-5 accuracy, model-size and FLOPs is also displayed in Fig. 5. The depth and total number of CONV layers for SqueezeNet and Inception are different as they consist of parallel CONV layers in their modules. The number of floating-point operations (FLOPs) represents how computationally expensive a CNN model is. The FLOPs of the model are computed using the TensorFlow built-in profiler. The top-5 accuracy is computed using the pre-trained models in Keras with the ILSVRC-2012 validation set, which consists of 50,000 images. The accuracy reported here is taken based on a single crop of the images. Accuracies reported in the literature usually use multiple-crops. However, for real-time operation, a single crop accuracy is regarded as more realistic. One can see that even though InceptionV3 has fewer number of parameters compared to ResNet, the number of FLOPs for the network is higher due to a greater number of CONV layers. All of these models are available pre-trained via Keras and can be extended to various applications by using transfer-learning [38].

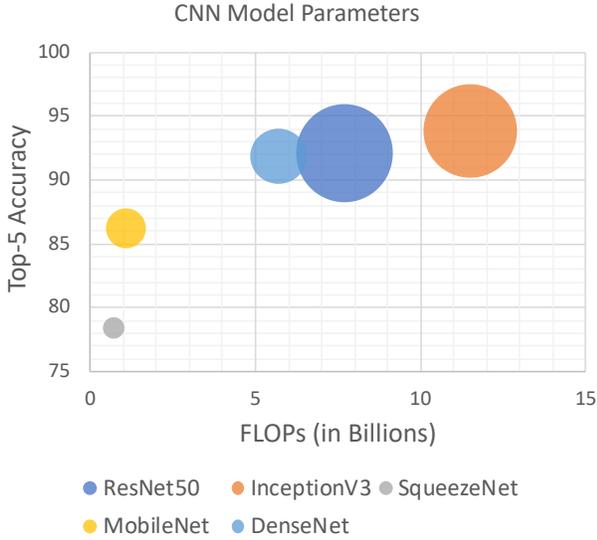

Fig. 5. Benchmarking metrics: Top-5 accuracy, FLOPs (in billions), and memory size of the DNN models; circle sizes represent model sizes in Table II.

### B. Multithreading

DNNs are computationally very expensive. For real-time operation on smartphones, executing the model on the main thread causes delay in capturing of frames and thus reduces the app throughput. This would also lead to a reduced number of FPS (frames per second). For applications where a DNN model can operate at slower rate than the frame rate of the app, multithreading needs to be adopted. Noting that multi-core processors are used in modern smartphones, a DNN model can be run on a secondary thread to create the needed computational bandwidth on the main thread to run the app at a desired FPS. This technique was used previously in [11] to allow a DNN model to run on a parallel thread by removing the computation burden from the main audio thread and thus preventing any audio frames to get skipped.

### C. Benchmarking Metrics

#### 1) Accuracy

The ILSVRC-2012 validation dataset is used to validate the CNN models. This validation set consists of 50,000 images of 1000 object categories. The DNN models are validated on the PC and smartphone platforms using the top-5 accuracy metric on a single crop. It should be noted that the accuracy reported in the literature involves multiple crops with an ensemble of classifiers which improves the accuracy. This is not possible for real-time apps running on smartphones. In other words, a single crop is more appropriate to consider when operating in real-time. For LeNet, the MNIST test set is used for validation. This set consists of 10,000 images of handwritten digits.

#### 2) CPU/GPU Consumption

For smartphone apps, CPU/GPU consumption is critical as this has a direct impact on the battery utilization. A higher consumption metric has a higher impact on the battery utilization. As TensorFlow for Mobile currently only supports running on the CPU, the CPU consumption of the Android app is measured here. The CoreML API utilizes the neural engine of iPhone 8, which in turn utilizes the GPU for the parallel computations of a CNN model. The GPU consumption of the iOS app is measured here as this is the way the majority of the computation is handled.

As far as LeNet is concerned, it is benchmarked as a non real-time app, or the CPU/GPU consumption is not monitored like the other real-time models.

#### 3) Real-Time Throughput

To evaluate the throughput of an app, the number of consecutive frames is displayed and measured per second on the screen (FPS). This is necessary for video-based apps that demand smooth visual perception of video data with an FPS of 24 or more.

The number of frames processed per second is also measured. This metric is highly dependent on the number of FLOPs in a model, as a higher number of FLOPs is directly proportional to a reduced number of frames processed per second and vice-versa. This metric is important to decipher which model is efficient to use. A model with a higher number of frames processed per second would be required for applications where throughput is critical. As can be seen in Fig. 5, models with high FLOPs have a higher accuracy. Therefore, in applications where accuracy is critical, number of frames processed per second can be reduced. When the model is run at frame rate, the FPS and frames processed per second are the same. When using multithreading, the number of frames processed per second is different than the FPS.

As LeNet is generally not used in real-time, the time taken per image to be classified is considered here as its throughput metric.

## IV. RESULTS AND DISCUSSION

Initially, LeNet was first implemented as Android and iOS apps using the developed approach. The accuracy of the validation set on the smartphone platforms was found to be 99.43%, the same as the PC platform, with a processing time of 5.66ms and 5.20ms per image on Pixel 2 and iPhone 8, respectively.

Then, the ILSVRC models were implemented as Android and iOS apps and the accuracy for the validation set consisting of 50,000 images of the ImageNet challenge dataset was examined. As illustrated in Fig. 6, the accuracy for the implemented apps was found to be practically the same, or within 0.5% difference. This slight difference is due to the fact that the precision of floating-point numbers is handled differently in ARM and Intel-based processors.

Next, to examine the CPU/GPU consumption and throughput, the models were run in real-time on the smartphones. For Android, the CPU consumption was computed using the Android Profiler [39] of the Android Studio IDE and the GPU consumption for iOS was computed using the

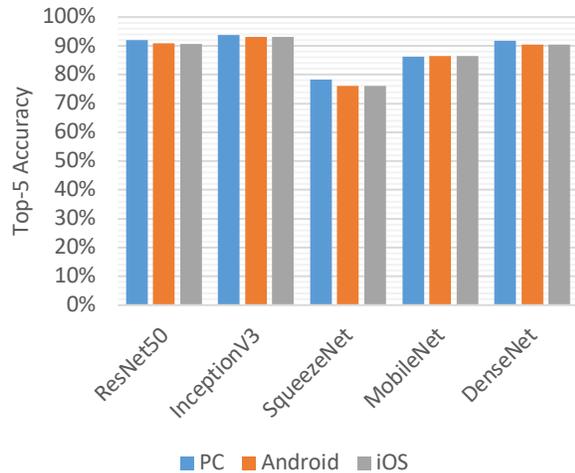

Fig. 6. Single crop top-5 accuracy of the ILSVRC challenge validation dataset on PC and smartphone platforms.

Instruments [40] performance analysis tool in the Xcode IDE. The FPS on Android was measured using the OpenCV camera API [41] and on iOS was measured using the Instruments performance analysis tool. The frames processed per second were measured for the multithreading approach by periodically computing it in the apps.

Fig. 7 provides the CPU/GPU consumption results that were obtained. From this figure, one can see that the best processing rate for Android was achieved by SqueezeNet which ran at 11 FPS. The CPU consumption of the Android apps was seen to be proportional to the number of FLOPs used by the apps. For iOS, only SqueezeNet was able to achieve greater than 24 FPS. As can be seen from this figure, the iOS apps benefited from access to the GPU leading to higher throughputs than their Android counterparts. The GPU consumption was seen to be proportional to the number of FLOPs used by the apps. It should be noted that a DNN app with a higher number of FLOPs drained the battery faster due to higher processor consumption.

When deploying multithreading, as illustrated in Fig. 8, the FPS remained constant at 30 FPS for all the Android apps but the number of frames processed per second was different for the apps. Multi-threading allowed running the GUI for a natural visual perception of 30 FPS while running the model in parallel at a lower rate. For the iOS versions of the apps, it can be seen that higher throughputs in both FPS and frames processed per second were achieved due to the use of the GPU. Obviously, the CPU and GPU consumptions increased when using multi-threading due to the use of concurrency while getting the benefit of being able to see video streams as they occurred.

A comparison of the apps implementing the DNN models indicates that SqueezeNet provides high energy efficiency as well as high throughputs on iOS smartphones but is not as accurate as the other models. It constitutes the model of choice for high throughput applications. MobileNets provides high accuracy in a multi-threaded setting at the expense of a lower energy efficiency and throughput. For applications where accuracy is the key requirement, one can use InceptionV3 in a multi-threaded setting as the model of choice. ResNet50 and DenseNet121 are also good choices providing relatively higher energy efficiency and throughput at the expense of 1-2% loss in accuracy.

## V. Conclusion and Future Extensions

This paper has presented in one place the steps needed in order to deploy deep learning inference networks on Android and iOS smartphones. It has been shown how to use the publicly available deep learning software tools to turn deep learning models into smartphone apps. In addition, it has been discussed how to enable real-time operation of such apps on smartphones. A benchmarking framework involving accuracy, CPU/GPU consumption, and real-time throughput has been devised to examine these models. The steps discussed have been validated using the benchmarking framework by considering six popular convolutional neural network models that are extensively used in deep learning applications. The benchmarking results have shown that the deep learning models are implemented without any significant loss in accuracy. It has also been shown that the use of multi-threading leads to achieving real-time throughputs. In summary, this paper has provided the guidelines and benchmarks for deploying deep learning inference models on smartphones as real-time apps.

It is worth mentioning here that the step-by-step guidelines provided in this paper can get extended through ONNX (Open Neural Network Exchange), which is a community project started by Facebook and Microsoft to provide a unified computational dataflow graph for deep neural networks. Such an extension would allow a model trained in ONNX supported frameworks to be used interchangeably. Currently, converters for CoreML and TensorFlow for other frameworks such as Caffe2, CNTK, PyTorch, etc., are being developed by ONNX, which would lead to seamless integration of models from different frameworks into the deployment approach presented in this paper.

Another extension involves the use of neural network compression methods. These methods involve reducing the size of models and thus the computation time by converting the weights of a model from floating-point numbers to integers with lower bits (quantization), by performing matrix decomposition, by pruning connections, etc. Once these methods are fully developed, supported by the smartphone hardware, and made available in the public domain, they can be incorporated seamlessly into the deployment approach presented in this paper.

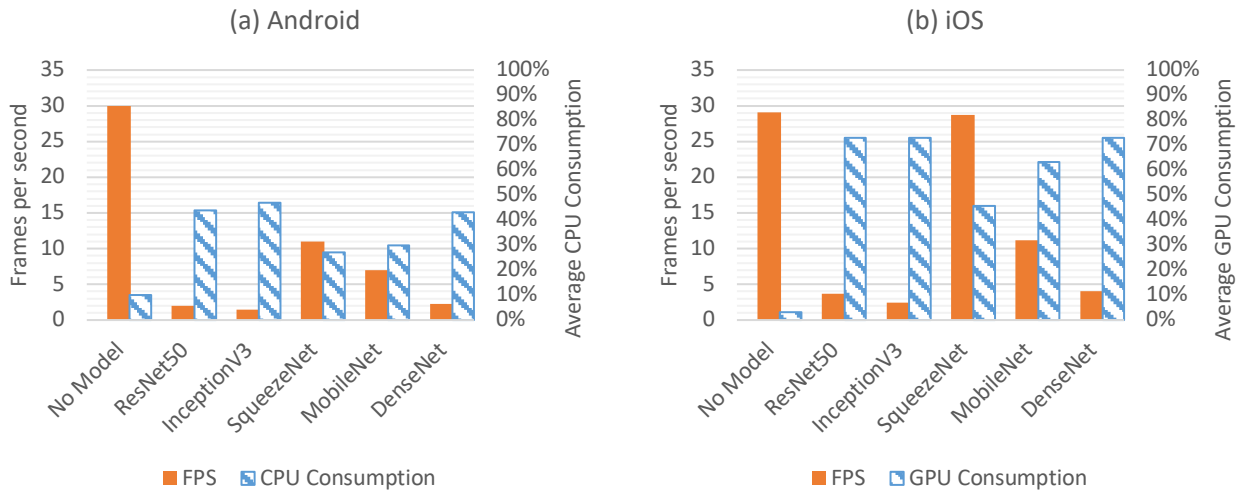

Fig. 7. (a) Average frames per second of the DNN Android apps are shown on the primary y-axis (left) and their average CPU consumption on the secondary y-axis (right), and (b) average frame per second of the DNN iOS apps are shown on the primary y-axis (left) and their average GPU consumption on the secondary y-axis (right). Note the number of frames processed per second is the same as FPS.

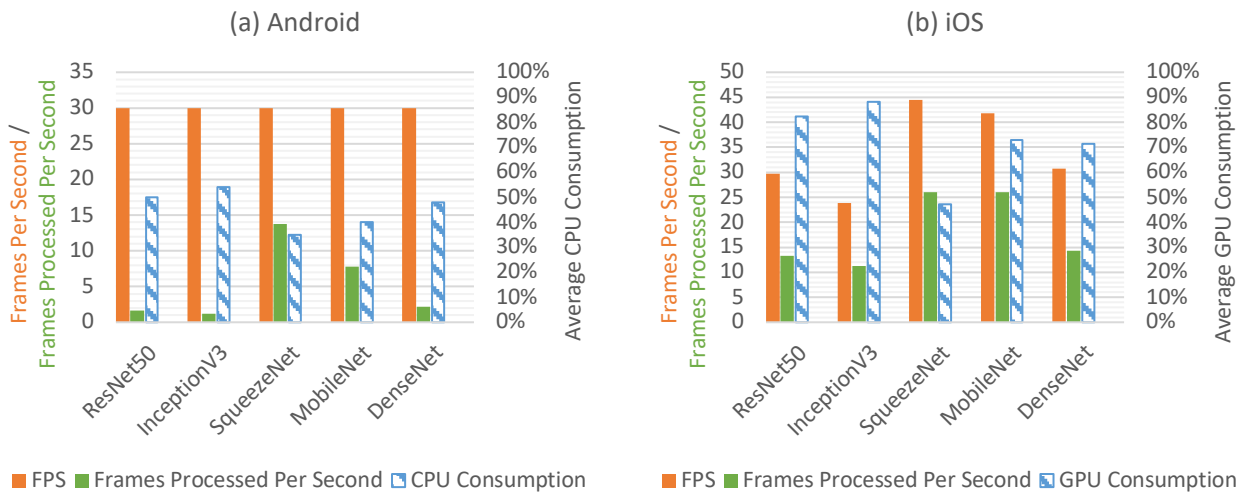

Fig. 8. (a) Average frames per second and frames processed per second of the DNN Android apps are shown on the primary y-axis (left) and their average CPU consumption on the secondary y-axis when using multithreading, and (b) average frames per second and frames processed per second of the DNN iOS apps are shown on the primary y-axis (left) and their average CPU consumption on the secondary y-axis (right) when using multithreading. Note the number of frames processed per second is different than FPS.
Jul-2018]